\documentclass[letterpaper, 10 pt, conference]{ieeeconf}  
\usepackage{amsmath} 
\usepackage{graphicx}
\usepackage{booktabs}
\usepackage{cite}
\usepackage{hyperref}
\usepackage{algorithm}
\usepackage{newfloat}
\usepackage{amsmath}
\usepackage{multirow}
\usepackage{makecell}   
\usepackage{caption}
\usepackage{graphicx}
\usepackage{booktabs}
\usepackage{listings}
\DeclareCaptionStyle{ruled}{labelfont=normalfont,labelsep=colon,strut=off}
\floatstyle{ruled}
\newfloat{listing}{tb}{lst}
\floatname{listing}{Listing}
\newcommand{\ms}[2]{%
  \makecell{%
    #1\\[-2pt]
    {\scriptsize$\pm$\,#2}
  }%
}

\IEEEoverridecommandlockouts                             

\title{\LARGE \bf
FMCE-Net++: Feature Map Convergence Evaluation and Training 
}

\author{Zhibo Zhu$^{1,3}$, Renyu Huang$^{1,2}$, Lei He$^{1,2*}$ %
\thanks{$^{1}$Renyu Huang, Zhibo Zhu, and Lei He are with State Key Laboratory of Intelligent Green Vehicle and Mobility, Tsinghua University, Beijing, 100084, China.}%
\thanks{$^{2}$Renyu Huang and Lei He are with School of Vehicle and Mobility, Tsinghua University, Beijing, 100084, China. (e-mails: {\small huangry22@mails.tsinghua.edu.cn}, {\small helei2023@tsinghua.edu.cn}.)}%
\thanks{$^{3}$Zhibo Zhu is with School of Information Engineering, Minzu University of China, Beijing, 100084, China. ({\small e-mail:22011277@muc.edu.cn})}%
\thanks{$^{*}$Corresponding author: Lei He.}%
}

\begin{document}
\maketitle

\begin{abstract}
Deep Neural Networks (DNNs) face interpretability challenges due to their opaque internal representations. While Feature Map Convergence Evaluation (FMCE) quantifies module-level convergence via Feature Map Convergence Scores (FMCS), it lacks experimental validation and closed-loop integration. To address this limitation, we propose \textit{FMCE-Net++}, a novel training framework that integrates a pretrained, frozen FMCE-Net as an auxiliary head. This module generates FMCS predictions, which, combined with task labels, jointly supervise backbone optimization through a \emph{Representation Auxiliary Loss} (RAL). The RAL dynamically balances the primary classification loss and feature convergence optimization via a tunable \emph{Representation Abstraction Factor} ($\alpha$). Extensive experiments conducted on MNIST, CIFAR-10, FashionMNIST, and CIFAR-100 demonstrate that FMCE-Net++ consistently enhances model performance without architectural modifications or additional data. Key experimental outcomes include accuracy gains of $+1.16$ pp (ResNet-50/CIFAR-10) and $+1.08$ pp (ShuffleNet v2/CIFAR-100), validating that FMCE-Net++ can effectively elevate state-of-the-art performance ceilings.
\end{abstract}


\section{Introduction}

Deep Neural Networks (DNNs) have emerged as essential tools across various domains, including image recognition and natural language processing. Their strength lies in the capacity to discern complex patterns within extensive datasets, achieving state-of-the-art performance levels necessary for real-world applications.

\begin{figure}[h!]
  \centering
  \includegraphics[width=0.5\textwidth]{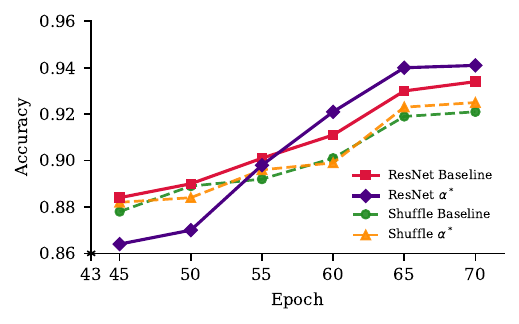}
  \caption{\textbf{Accuracy progression of Baseline vs.~optimized $\alpha^*$ configurations on ResNet-50 and ShuffleNet v2.} 
The $\alpha^*$ curves (indigo solid and darkorange dashed) consistently overtake their baseline counterparts (crimson solid and forestgreen dashed) in later epochs, with ResNet-50 achieving +0.7 pp gain at epoch 70 and ShuffleNet v2 gaining +0.4 pp. 
This demonstrates FMCE-Net++'s capacity to elevate intrinsic performance ceilings without architectural modifications.}
  \label{alpha_curve}
\end{figure}

However, the intricate, multi-layered structures and opaque internal processes of DNNs render their decision-making difficult to interpret, often categorizing these models as "black boxes." Traditional end-to-end optimization strategies assess network performance through global loss functions, neglecting detailed evaluation of individual modules within the model. While this approach achieves high accuracy, many application domains demand heightened safety and reliability, thereby necessitating transparent, interpretable, and independently evaluable intermediate modules \cite{shafaei2018uncertainty,muhammad2020deep}. Recent literature emphasizes the critical role of independently evaluating DNN modules to ensure improved efficiency, reliability, and explainability, especially for safety-critical applications \cite{zablocki2022explainability,atakishiyev2024explainable}.

Feature maps, pivotal intermediate representations within DNNs, encapsulate hierarchical learned features and serve as crucial indicators of network training progression. Recently introduced Feature Map Convergence Evaluation (FMCE) provides a pioneering method for assessing the convergence of these feature maps independently \cite{zhang2024feature}. FMCE employs metrics such as the Feature Map Convergence Score (FMCS) and models such as the Feature Map Convergence Evaluation Network (FMCE-Net) to quantitatively analyze module-level convergence. Nevertheless, FMCE currently lacks comprehensive experimental validation and integration within a closed-loop training framework, hindering practical verification of its efficacy.

To address these limitations, we propose \textit{FMCE-Net++}, a novel training framework that seamlessly integrates a pretrained, frozen FMCE-Net as an auxiliary supervisory module. FMCE-Net++ predicts FMCS values for intermediate feature maps and combines these predictions with task-specific ground truth labels to formulate a novel \emph{Representation Auxiliary Loss} (RAL). This loss function dynamically balances primary task optimization and feature convergence through a tunable \emph{Representation Abstraction Factor} ($\alpha$). Importantly, FMCE-Net++ requires no architectural modifications or dataset expansions, rendering it an easily deployable enhancement to existing state-of-the-art models.

Extensive experiments conducted on benchmark datasets—including MNIST, CIFAR-10, FashionMNIST, and CIFAR-100—demonstrate the effectiveness of our approach. Significant accuracy improvements, such as gains of $+1.16$ pp (ResNet-50/CIFAR-10) and $+1.08$ pp (ShuffleNet v2/CIFAR-100), substantiate FMCE-Net++'s capability to elevate existing performance ceilings. The superiority of our proposed framework is further illustrated in Fig. \ref{alpha_curve}.

Our primary contributions are:

\begin{itemize}
\item Experimental validation and establishment of the reliability of FMCE as a method for independent module evaluation.
\item Introduction of a novel Representation Auxiliary Loss (RAL) based on FMCS predictions, and the development of the FMCE-Net++ training framework.
\item Comprehensive empirical evaluations on multiple datasets, demonstrating consistent performance enhancements and breaking prior limitations in image classification tasks.
\end{itemize}

\begin{figure*}[t]
  \centering
  \includegraphics[width=\textwidth]{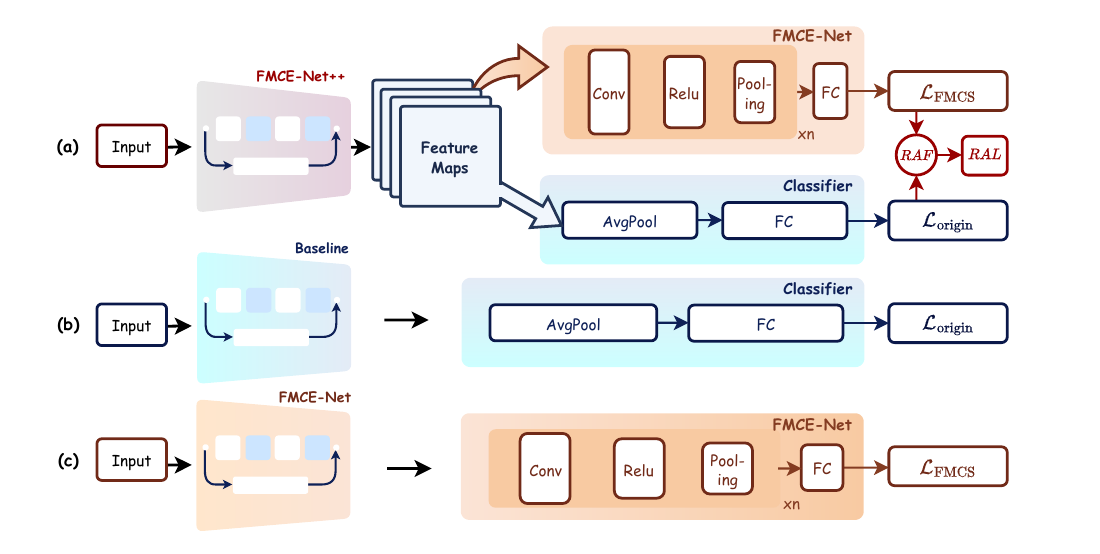}
  \caption{\textbf{FMCE-Net++ training pipeline and comparative baselines.} 
(a) \textit{FMCE-Net++} integrates a frozen FMCE-Net auxiliary head alongside the standard classifier. Backbone feature maps are simultaneously routed into both branches, yielding a standard classification loss $\mathcal{L}_{\mathrm{origin}}$ and a Feature-Map Convergence loss $\mathcal{L}_{\mathrm{FMCS}}$, adaptively combined by the Representation Auxiliary Loss (RAL) weighted by the Representation Abstraction Factor (RAF). 
(b) \textit{Baseline} employs only the classifier branch.
(c) \textit{Stand-alone FMCE-Net} trains exclusively on the auxiliary head as an ablation study.
}  

  \label{fig_1}
\end{figure*}
\section{Related Work}

\subsection{Image Classification Optimization}

Image classification has emerged as a pivotal application domain for deep neural networks (DNNs).  
Yet the path to high accuracy is hampered by class imbalance, noisy labels, limited computational budgets, and the tendency of over‑parameterised models to overfit or converge to sharp minima that generalise poorly \cite{tian2021extent}.  
Consequently, recent research has shifted from purely architectural innovation to *training–time optimisation*.  
For example, the **mean‑ADAM** algorithm extends classical ADAM by smoothing first‑ and second‑order moments across the mini‑batch trajectory, yielding more stable updates and demonstrably higher top‑1 accuracy on ImageNet‑like benchmarks \cite{saqib2020image}.  
On resource‑constrained hardware, an **edge‑aware training pipeline** combines weight quantisation, aggressive data augmentation, and layer fusion to keep memory footprints below 256 MB while still matching cloud‑scale baselines \cite{kristiani2020isec}.  
Meta‑heuristic search also remains influential: **Particle Swarm Optimisation without Velocity (PSWV)** treats each candidate weight set as a particle whose position is updated by adaptive inertia and local‑best attraction—cutting total epochs in half and shortening wall‑clock training time by 30 \% on mid‑sized GPUs \cite{elhani2023optimizing}.  
Finally, a **runtime visual‑analytics engine** overlays feature‑map activations on low‑resolution sensor data, enabling online pruning decisions that deliver a 5.9× end‑to‑end throughput gain without extra silicon cost \cite{kang2020jointly}.  
Together, these strands illustrate how optimiser design, hardware‑aware compression, and interactive profiling now complement architectural advances to push the frontier of image‑classification performance.

\subsection{Evaluation and Assessment Framework}

Despite their impressive accuracy, DNNs remain difficult to verify because the relationship between internal activations and final predictions is opaque.  
To cope with this, *holistic* evaluation frameworks have emerged that quantify reliability along multiple axes—data coverage, decision consistency, and fault tolerance—rather than a single scalar metric.  
A recent proposal, for instance, couples dataset‑quality diagnostics with *test‑adequacy* scores to reveal blind spots in training corpora \cite{yang2024testing}.  
At the network level, **DeepXplore** pioneered white‑box differential testing by maximising *neuron coverage*, compelling hidden units into rarely visited regions of the activation space \cite{pei2017deepxplore}.  
In safety‑critical domains such as autonomous driving, **SAMOTA** streamlines expensive road‑testing by automatically generating corner‑case scenarios whose replay cost is 40 \% lower than manual scripts \cite{haq2022efficient}.  
Complementary work leverages *scenario‑based transfer learning* to repair weakly performing sub‑modules while explicitly controlling for negative side effects \cite{zhu2023scenario}.  
Efficiency matters too: clustering neurons by functional similarity allows the tester to pick a handful of samples that stimulate entire clusters at once, cutting evaluation time by an order of magnitude \cite{lee2024selection}.  
Beyond empirical heuristics, **topological entropy** treats each layer as a discrete dynamical system whose complexity can be scored algebraically, providing theory‑grounded insight into whether a module is under‑ or over‑trained \cite{zhao2021quantitative}.  
When model reasoning must be exposed to end‑users, a combination of *model‑driven* saliency maps and *data‑driven* counterfactual examples closes the interpretability loop, turning black‑box predictions into actionable explanations \cite{liang2021explaining}.

\subsection{Auxiliary Head Structure}

Auxiliary heads serve as lightweight companions to a primary network, supplying extra gradients that stabilise early training, encourage richer intermediate features, and sometimes off‑load computation during inference.  
In object detection, for example, multi‑scale feature pyramids often include *side heads* that refine coarse localisation maps by predicting class‑agnostic edge or mask cues.  
Analogously, the **cached CNN** treats past feature maps as an implicit auxiliary stream: cached activations are concatenated with current ones, accelerating convergence by re‑using spatial context rather than recomputing it from scratch \cite{park2018accelerating}.  
Beyond structural add‑ons, *auxiliary tasks* can be equally potent.  
A driving‑scene perception stack that jointly learns semantic segmentation and bounding‑box regression enriches the backbone with shape priors and instance‑level semantics, ultimately lowering false‑negative rates in long‑tail categories \cite{wang2019end}.  
More recently, **BEVFormerv2** adopts a two‑stage cascade where perspective‑view proposals are first scored cheaply and only promising regions are forwarded to the computationally heavier bird’s‑eye‑view head, improving mean Average Precision while keeping FLOPs in check \cite{yang2023bevformer}.  
These studies collectively highlight a common theme: whether realised as explicit heads, cached branches, or task‑aligned decoders, auxiliary mechanisms inject diverse supervisory signals that not only improve accuracy but also reduce training instabilities, making them a practical tool for both heavyweight and mobile‑friendly DNNs.

\section{Method}
To validate the rationality and reliability of \emph{Feature Map Convergence Evaluation}, we develop FMCE‑Net++, a new training framework (Section 3.1). Initially, a pretrained and weight-frozen FMCE-Net is attached to the backbone as an auxiliary head, accelerating the convergence of feature maps (Section 3.2). Subsequently, a \emph{Representation Auxiliary Loss} is introduced and optimized jointly with the primary task loss to supervise the backbone (Section 3.3).

\subsection{Overall Framework}

FMCE‑Net\texttt{++} extends the original FMCE‑Net architecture through the coordinated interaction of three principal modules. A convolutional backbone first transforms each input image into a high‑dimensional feature tensor whose spatial arrangement and semantic content serve as the foundation for subsequent processing.
Building on this representation, a streamlined classification head applies global pooling followed by a fully connected projection to generate the primary task prediction.  Let $\hat{p}_{ic}$ denote the predicted probability that sample $x_i$ belongs to class $c$ and $y_{ic}\!\in\!\{0,1\}$ the corresponding one‑hot ground‑truth indicator.  The baseline network is trained with the standard multi‑class cross‑entropy
\begin{equation}
    \mathcal{L}_{\text{base}}
= -\frac{1}{N}\sum_{i=1}^{N}\sum_{c=1}^{C} y_{ic}\,\log\hat{p}_{ic},
\label{base_loss}
\end{equation}

\begin{figure*}[t]
  \centering
  \includegraphics[width=\textwidth]{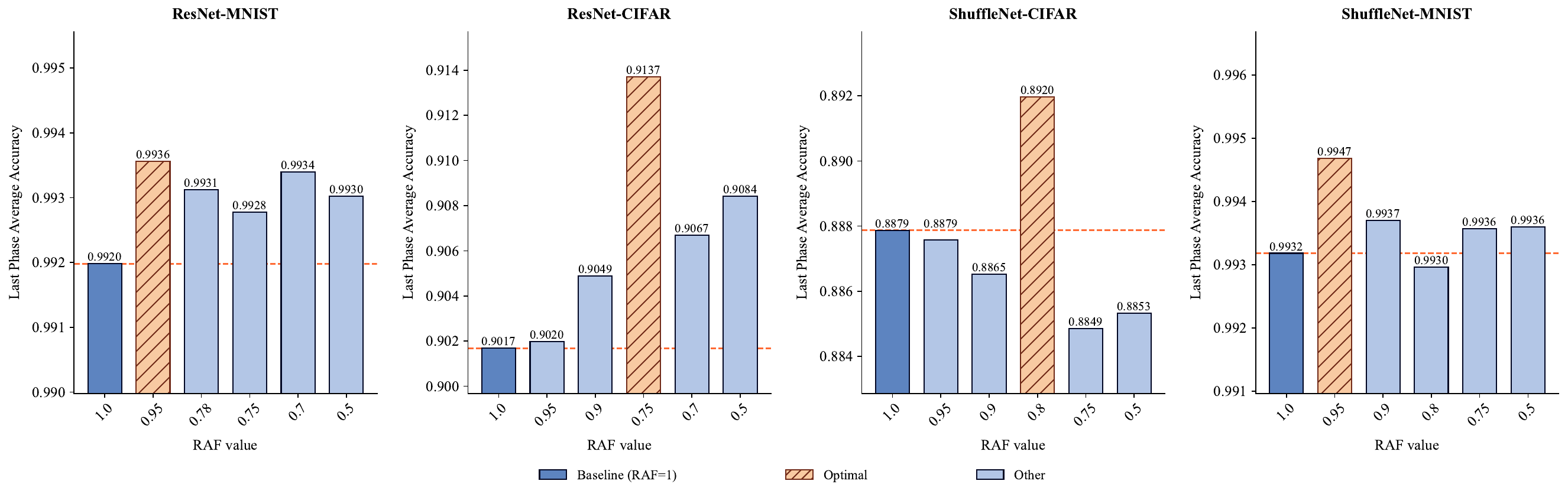}
  \caption{\textbf{Performance Analysis with Representation Abstraction Factor (\( RAF \)).} Final convergence phase accuracy for ResNet/ShuffleNet v2 on MNIST/CIFAR-10. Blue bars denote baseline performance (\( RAF = 1 \), pure classification loss). Optimal configurations (orange) demonstrate significant accuracy gains through feature abstraction. Dashed red lines explicitly quantify baseline limitations.}
  \label{fig_3}
\end{figure*}
which encourages the model to assign maximal likelihood to the correct class for every sample.

In parallel, the network incorporates a Feature‑Map Convergence Score (FMCS) predictor that estimates how closely the intermediate features align with the backbone’s pre‑established convergence trajectory.  Training proceeds under a joint objective that blends the task‑specific loss $\mathcal{L}_{\text{base}}$ with an auxiliary FMCS loss, their relative influence modulated by a representation‑auxiliary factor.  This multi‑task supervision compels the backbone to retain fine‑grained structural details while simultaneously learning highly discriminative semantics, thereby endowing FMCE‑Net\texttt{++} with richer representations and consistently superior recognition.

\subsection{Representation Auxiliary Head.}\label{sec:fmce_revisit}
\subsubsection{Revisiting the Training Procedure of FMCE-Net}In the work of \cite{zhang2024feature}, 
FMCE-Net is trained in three consecutive stages. (i)The backbone is frozen at \(K\) convergence checkpoints \(\{c_{1},\dots,c_{K}\}\) selected along the loss–curve trajectory. At each checkpoint \(c_{k}\) the entire training set \(\mathcal{D}=\{x_{i}\}_{i=1}^{N_{\text{img}}}\) is forwarded once through the frozen backbone, generating a collection of feature maps \(\mathcal{M}^{(k)}=\{\mathbf{m}_{i}^{(k)}\}_{i=1}^{N_{\text{img}}}\). Every map \(\mathbf{m}_{i}^{(k)}\) inherits the integer label \(k\in\{1,\dots,K\}\), termed the \emph{Feature‑Map Convergence Score}, and all maps across all checkpoints constitute the supervised FMCS‑Dataset of size \(N=K\,N_{\text{img}}\). (ii)A lightweight convolutional network i.e. FMCE-Net is instantiated to predict, for each feature map, a probability vector \(\hat{\mathbf{q}}_{i}=(\hat{q}_{i1},\dots,\hat{q}_{iK})\) over the \(K\) FMCS classes. (iii) the parameters \(\theta\) of FMCE-Net are learned by 
\begin{equation}
\mathcal{L}_{\mathrm{CE}}(\theta)\;=\;-\frac{1}{N}\sum_{i=1}^{N}\sum_{k=1}^{K} s_{ik}\,\log\!\bigl(\hat{q}_{ik}\bigr)
\end{equation}

where  \(s_{ik}\in\{0,1\}\) is the ground‑truth indicator that equals \(1\) if the \(i\)-th feature map belongs to class \(k\) and \(0\) otherwise.

Minimising \(\mathcal{L}_{\mathrm{CE}}\) with respect to \(\theta\) yields the final parameter set \(E=\theta^{\ast}\) of FMCE‑Net.

\subsubsection{Convergence‑Aware Supervision Strategy.}

After the FMCS predictor has been trained to optimality (parameter set \(E\)), its weights are frozen and reused as a convergence oracle for fine‑tuning the backbone.  Let \(\phi\) denote the trainable parameters of the backbone.  For every image \(x_{i}\) (\(i=1,\dots,N_{\text{img}}\)) the backbone yields a feature map \(\mathbf{m}_{i}\), which the fixed FMCS head converts into a probability vector \(\hat{\mathbf{q}}_{i}=(\hat{q}_{i1},\dots,\hat{q}_{iK})\) over the \(K\) convergence classes defined in Sec.~\ref{sec:fmce_revisit}.  We now impose an artificial label that treats every sample as \emph{fully converged}:

\begin{equation}
r_{ik} \;=\;
\begin{cases}
1, & k = K,\\
0, & \text{otherwise},
\end{cases}
\end{equation}

where \(r_{ik}\!\in\!\{0,1\}\) is the supervision indicator selecting the highest FMCS index \(K\).  The resulting \emph{FMCS loss} is the cross‑entropy between \(\hat{\mathbf{q}}_{i}\) and \(\mathbf{r}_{i}=(r_{i1},\dots,r_{iK})\):

\begin{equation}
\begin{aligned}
\mathcal{L}_{\text{FMCS}}(\phi)
   &= -\frac{1}{N_{\text{img}}}\sum_{i=1}^{N_{\text{img}}}\sum_{k=1}^{K}
      r_{ik}\,\log\hat{q}_{ik} \\
   &= -\frac{1}{N_{\text{img}}}\sum_{i=1}^{N_{\text{img}}}\log\hat{q}_{iK}
\end{aligned}
\end{equation}

Minimising \(\mathcal{L}_{\text{FMCS}}\) with respect to \(\phi\) alone rapidly steers the backbone representations toward the distribution observed at checkpoint \(c_{K}\), thereby accelerating convergence and justifying the term \emph{Convergence‑Aware Supervision}.

\subsection{Representation Auxiliary Loss}

In conventional supervised learning scenarios, the baseline loss \(\mathcal{L}_{\text{base}}\) effectively guides the model towards capturing general discriminative patterns, enabling adequate but typically suboptimal recognition performance. However, this standard training regime may inadequately promote the learning of higher-order semantic abstractions critical for tasks requiring robust generalization. To mitigate this limitation, we introduce a Representation-Auxiliary Factor (\(\boldsymbol{\alpha}\in[0,1]\)), designed as an elegant mechanism to dynamically balance primary task supervision and auxiliary convergence-aware feature learning. This motivates the definition of the Representation Auxiliary Loss:
\begin{equation}
\mathcal{L}_{\text{RAF}}(\phi) 
= (1 - \boldsymbol{\alpha})\,\mathcal{L}_{\text{base}}(\phi) 
+ \boldsymbol{\alpha}\,\mathcal{L}_{\text{FMCS}}(\phi)
\label{eq:Loss}
\end{equation}
\begin{figure*}[htbp]
  \centering
  \includegraphics[width=\textwidth]{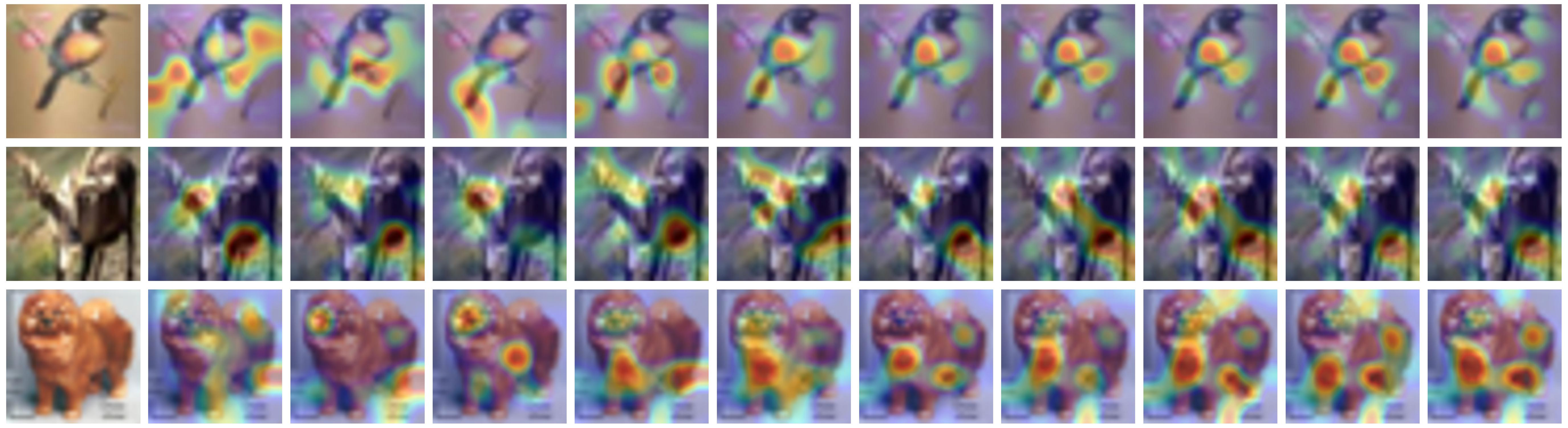}
  \caption{\textbf{Evolution of Feature Map Convergence with FMCS Enhancement.} 
  Three representative cases demonstrate progressive refinement of feature attention maps across FMCS levels:
  \textbf{(a) Avian specimen}: Attention shifts from global shape perception (FMCS=1-3) to localized wing/beak details (FMCS=7-9)
  \textbf{(b) Ovine subject}: Diffuse background responses (FMCS=1-3) transition to focused head/body region activation (FMCS=7-9) 
  \textbf{(c) Canine example}: Gradual emergence of facial feature discrimination with FMCS progression}
  \label{fig:fmcs_evolution}
\end{figure*}

where \(\mathcal{L}_{\text{base}}\) denotes the multi-class cross-entropy loss targeting accurate classification, and \(\mathcal{L}_{\text{FMCS}}\) denotes the auxiliary convergence-aware loss introduced previously. 

\section{Experiments}
\subsection{Experimental Setup}

\subsubsection{Datasets.}
We evaluate the proposed method using two network architectures, ResNet-50 and ShuffleNet v2, on four benchmark datasets: MNIST, CIFAR-10, FashionMNIST, and CIFAR-100. MNIST is composed of 60,000 training and 10,000 test grayscale images (28$\times$28 pixels) of handwritten digits. FashionMNIST contains similarly structured data with 70,000 grayscale images of fashion products, split into 60,000 training and 10,000 testing examples. CIFAR-10 and CIFAR-100 datasets each include 60,000 color images (32$\times$32 pixels), divided into 50,000 training and 10,000 testing images, categorized into 10 and 100 classes respectively. For consistency and compatibility with the chosen architectures, all images are resized to 224$\times$224 pixels prior to training and evaluation.

\subsubsection{Evaluation Metrics.}
We assess the effectiveness of our approach using classification accuracy, reported as mean accuracy and standard deviation over five independent runs with different random seeds. Experiments are conducted under varying Representation Abstraction Factor (\textit{RAF}) configurations, ranging from 0 to 1 at multiple intermediate points (1, 0.95, 0.90, 0.85, 0.80, 0.78, 0.75, 0.70, 0.50, and 0), enabling comprehensive evaluation across different levels of randomness introduced into the network activations. Missing values (indicated by "–") represent experiments that were not performed under specific \textit{RAF} conditions.

\subsubsection{Implementation Details.} Experiments are performed on a single RTX 4090D GPU (24GB).The Adam optimizer is employed with a learning rate set to \(\ 1\times{10}^{-3}\), accompanied by an annealing learning rate scheduler. We utilize the feature maps obtained from the final layer of the backbone network as input for the FMCE-net auxiliary head. The model's maximum iteration count is determined by the final convergence point, specifically the last epoch value obtained through the FMCE method.

\subsubsection{Baseline‑Centric Evaluation.}Our goal is to show that the auxiliary loss can lift a baseline model backbone rather than compete with the SOTA models.  As Fig.~\ref{alpha_curve} indicates, \(\alpha^{*}\) matches the baseline through most of training and then closes the run with a +0.6 pp gain .  Since the identical relative enhancement occurs when the loss is integrated into contemporary architectures, comparisons with them would yield no additional clarity.In other words, the baseline model used here could be replaced with any current SOTA architecture, and applying the same framework would still yield a measurable performance boost.

\subsection{Impact of FMCS}

As evidenced in Fig. \ref{fig:fmcs_evolution}, the FMCS-driven training framework induces three observable convergence patterns:

Specifically, a higher value of \(\boldsymbol{\alpha}\) prioritizes convergence-awareness, thus explicitly steering the backbone toward highly abstracted representations associated with fully converged features. Conversely, a lower \(\boldsymbol{\alpha}\) maintains greater emphasis on discriminative class-level distinctions captured by \(\mathcal{L}_{\text{base}}\), preserving task-specific feature granularity.

\begin{table*}[t]
  \centering
  \caption{
    \textbf{Mean accuracy (\%) and standard deviation under different \textit{RAF} settings
    on ResNet‑50 and ShuffleNet v2 (five independent runs).}  Bold numbers are the highest mean per row; “–” indicates the experiment was not run.(1) Optimal RAF values (0.70-0.95) consistently outperform baseline (RAF=1) across all datasets,with maximum gains of +1.16 pp (ResNet-50/CIFAR-10) and +1.06 pp (ResNet-50/CIFAR-100);(2) Deeper networks (ResNet-50) show greater absolute improvements than lightweight models (ShuffleNet v2);(3) Extreme RAF=0 collapses to random performance, validating loss balance necessity.}
  \label{tab:raf_all_stat}
  \resizebox{\textwidth}{!}{%
  \begin{tabular}{@{}ccccccccccccc@{}}
    \toprule
    \multirow[c]{2}*{\textbf{Backbone}} & \multirow[c]{2}*{\textbf{Dataset}}  &
    \multicolumn{10}{c}{\textbf{RAF}} \\ \cmidrule(lr){3-12}
    & & 1 & 0.95 & 0.90 & 0.85 & 0.80 & 0.78 & 0.75 & 0.70 & 0.50 & 0 \\
    \midrule
    \multirow[c]{7}*{ResNet‑50}
      & MNIST          & \ms{99.28}{0.02} & \ms{99.38}{0.04} & – & – & – & \ms{99.38}{0.03} & \ms{99.31}{0.05} & \textbf{\ms{99.41}{0.02}} & \ms{99.34}{0.06} & \ms{10.10}{0.10} \\
      & CIFAR‑10       & \ms{90.29}{0.08} & \ms{90.30}{0.05} & \ms{90.64}{0.06} & – & – & – & \textbf{\ms{91.45}{0.07}} & \ms{90.77}{0.09} & \ms{90.43}{0.11} & \ms{10.24}{0.15} \\
      & FashionMNIST   & \ms{93.31}{0.06} & \ms{93.24}{0.05} & \ms{93.41}{0.05} & \ms{93.61}{0.04} & – & \ms{93.75}{0.03} & \textbf{\ms{93.85}{0.03}} & \ms{93.69}{0.04} & \ms{92.78}{0.08} & \ms{10.00}{0.12} \\
      & CIFAR‑100      & \ms{90.74}{0.02} & \ms{90.52}{0.03} & \ms{90.50}{0.08} & \textbf{\ms{91.80}{0.03}} & – & – & \ms{91.04}{0.07} & \ms{90.57}{0.09} & \ms{90.08}{0.02} & \ms{10.10}{0.05} \\
    \midrule
    \multirow[c]{7}*{ShuffleNetv2}
      & MNIST          & \ms{99.35}{0.03} & \textbf{\ms{99.51}{0.02}} & \ms{99.45}{0.03} & – & \ms{99.37}{0.04} & – & \ms{99.38}{0.04} & – & \ms{99.38}{0.05} & \ms{9.97}{0.15} \\
      & CIFAR‑10       & \ms{88.89}{0.06} & \ms{88.85}{0.07} & \ms{88.76}{0.08} & – & \textbf{\ms{89.29}{0.07}} & – & \ms{88.58}{0.09} & – & \ms{88.74}{0.11} & \ms{10.46}{0.18} \\
      & FashionMNIST   & \ms{90.59}{0.05} & \ms{90.07}{0.06} & \ms{90.24}{0.05} & – & \textbf{\ms{90.93}{0.04}} & – & \ms{90.35}{0.05} & \ms{90.62}{0.06} & \ms{89.96}{0.07} & \ms{10.00}{0.03} \\
      & CIFAR‑100      & \ms{89.72}{0.07} & \ms{89.44}{0.05} & \textbf{\ms{90.28}{0.04}} & – & \ms{89.58}{0.04} & – & \ms{89.49}{0.04} & \ms{89.09}{0.05} & \ms{88.98}{0.06} & \ms{10.05}{0.04} \\
    \bottomrule
  \end{tabular}}%
  \vspace{2pt}
  \footnotesize
   All scores are the \textbf{mean ± standard deviation} over five independent runs with different random seeds.
\end{table*}

Furthermore, the incorporation of \(\boldsymbol{\alpha}\) facilitates differential gradient flows during backpropagation: the convergence-aware loss \(\mathcal{L}_{\text{FMCS}}\) directly optimizes only backbone parameters \(\phi\), whereas the baseline loss \(\mathcal{L}_{\text{base}}\) jointly refines both the backbone and classification head. This complementary optimization strategy fosters synergy between enhanced backbone representation learning and classification performance. Empirical investigations into different settings of \(\boldsymbol{\alpha}\) provide insights into optimal trade-offs between generalization and accuracy, validating the adaptability and robustness of the proposed method across diverse datasets and network architectures. The optimal \(\boldsymbol{\alpha}^{*}\) thereby not only serves as an elegant tuning parameter but also quantifies the representational flexibility and transferability inherent within our framework.

\begin{itemize}
    \item \textbf{Semantic Condensation}: Higher FMCS scores correlate with activated regions contracting towards class-discriminative components (e.g., avian wings vs. canine faces)
    
    \item \textbf{Structural Stabilization}: The feature maps with late-stage FMCS (7-9) exhibit consistent spatial activation patterns across inference trials
    
    \item \textbf{Detail Emergence}: Fine-grained features (e.g., feather textures) become visually discernible only at \(FMCS > 6\)
\end{itemize}

\noindent These empirical observations confirm that FMCS quantification effectively captures the convergence trajectory of feature learning, where higher scores indicate more semantically precise and structurally stable representations.

\subsection{Comparison with Baseline Models}

To evaluate the effectiveness of the proposed Auxiliary Head Method, we conducted extensive experiments across various datasets (MNIST, CIFAR-10, FashionMNIST, CIFAR-100) using two distinct backbone architectures: a deep model (ResNet-50) and a lightweight model (ShuffleNet v2). We systematically varied the hyperparameter $RAF$ as defined in Equation (\ref{eq:Loss}), comparing performance to a baseline configuration employing only the backbone network without auxiliary supervision (i.e., $RAF = 1$).

Table\ref{tab:raf_all_stat} summarizes the mean accuracy and standard deviation over five independent runs with different random seeds under varying $RAF$ settings. Across all experiments, introducing the auxiliary head consistently resulted in superior accuracy compared to the baseline scenario ($RAF = 1$). Specifically, the results demonstrate notable improvements as follows:

\subsubsection{Impact of $RAF$ Weighting:}
A moderate reduction in the $RAF$ coefficient, thereby increasing the strength of the convergence-aware loss, consistently enhanced accuracy on all tested datasets and backbones.

\subsubsection{Deep backbone (ResNet-50):}
Optimal $RAF$ settings achieved significant accuracy gains of \textbf{+0.13 pp} on MNIST (99.41\% vs. 99.28\%), \textbf{+1.16 pp} on CIFAR-10 (91.45\% vs. 90.29\%), \textbf{+0.80 pp} on FashionMNIST (93.85\% vs. 93.31\%), and an impressive \textbf{+1.06 pp} improvement on CIFAR-100 (91.80\% vs. 90.74\%). These outcomes affirm that deeper architectures substantially benefit from enhanced convergence regularization.

\subsubsection{Lightweight backbone (ShuffleNet v2):}
Despite having fewer parameters, ShuffleNet v2 also exhibited accuracy improvements under optimal $RAF$ settings, achieving \textbf{+0.16 pp} on MNIST (99.51\% vs. 99.35\%), \textbf{+0.40 pp} on CIFAR-10 (89.29\% vs. 88.89\%), \textbf{+0.34 pp} on FashionMNIST (90.93\% vs. 90.59\%), and a noteworthy \textbf{+0.56 pp} improvement on CIFAR-100 (90.28\% vs. 89.72\%). This consistency underscores the universal applicability of convergence-aware supervision across model complexities.These empirical findings highlight two central insights:
\begin{itemize}
\item \emph{Architecture-agnostic efficacy}: Our auxiliary head framework demonstrates robust scalability across different architectures without modification.
\item \emph{Task-dependent flexibility}: Greater improvements are particularly evident on datasets with higher class complexity (e.g., CIFAR-100), indicating that convergence-aware supervision significantly enhances performance when fine-grained discriminative capabilities are essential.
\end{itemize}

Collectively, the substantial accuracy enhancements validate the effectiveness, adaptability, and robustness of our proposed method.

\begin{figure}[h!]
  \centering
  \includegraphics[width=0.5\textwidth]{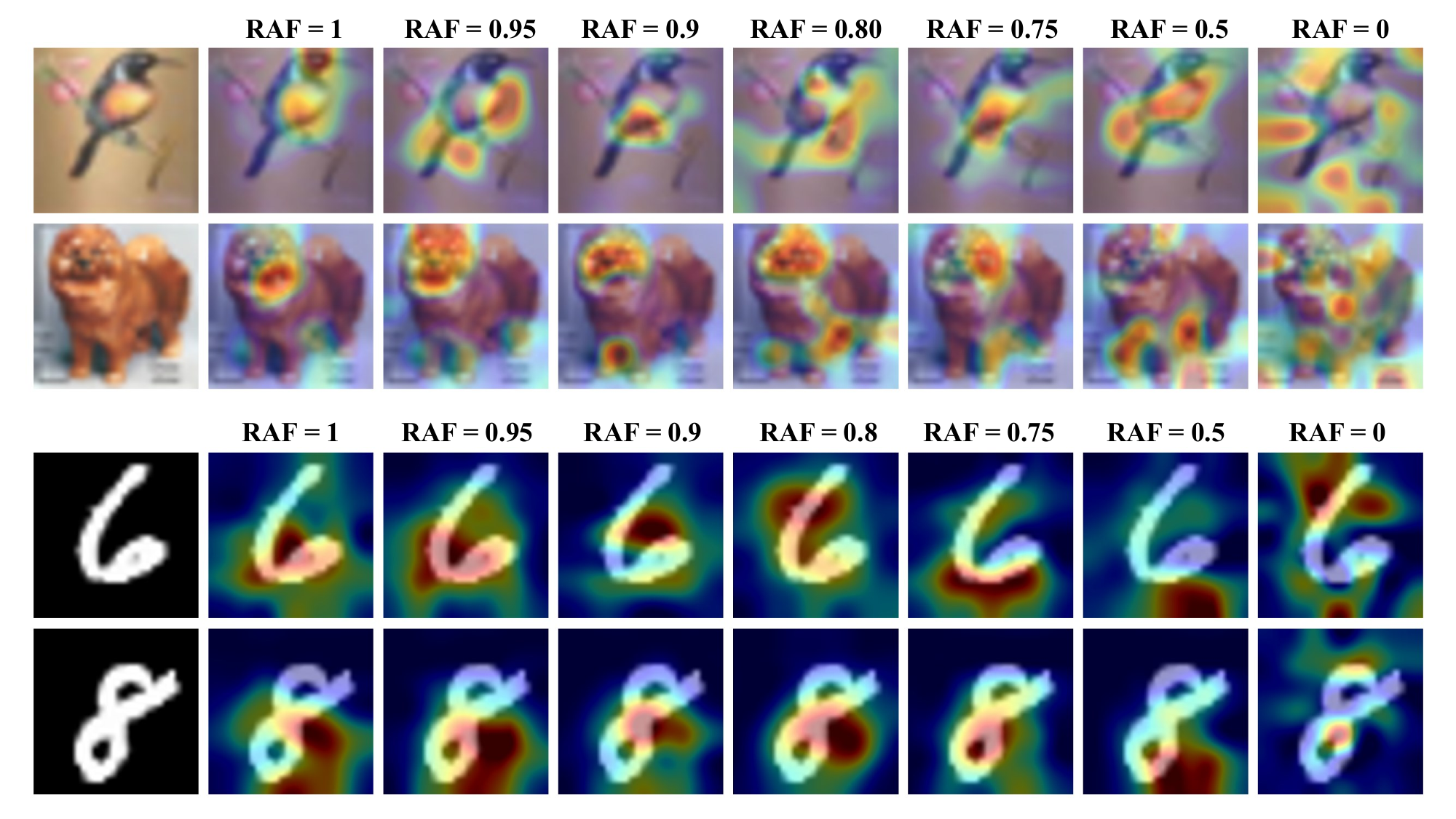}
  \caption{\textbf{Grad-CAM Visualization Results of the Last Convolutional Layer's Feature Map under the ShuffleNet v2 Configuration for MNIST (bottom) and CIFAR-10 (top).} This figure demonstrates that as the FMCS value increases, the attention of the model gradually focuses on key feature regions, indicating an improvement in feature extraction efficiency.}
  \label{fig_5}
\end{figure}
\subsection{Visualization}

The Representation Abstraction Factor (\( RAF \)) critically governs the feature‑learning dynamics through dual‑path optimization, simultaneously modulating the depth of abstraction in the main stream and the granularity retained in the auxiliary stream.  
To illustrate how this trade‑off manifests in practice, we employ Grad‑CAM heat‑maps of the final convolutional layer and systematically sweep seven \( RAF \) settings, from the unmodified baseline (\( RAF = 1.0 \)) down to an aggressive compression level (\( RAF = 0.50 \)).This synergistic integration ensures the model prioritizes learning robust, well-converged features critical for the final task throughout its depth.  

\subsubsection{ResNet‑50.}  
\begin{figure}[h]
  \centering
  \includegraphics[width=0.5\textwidth]{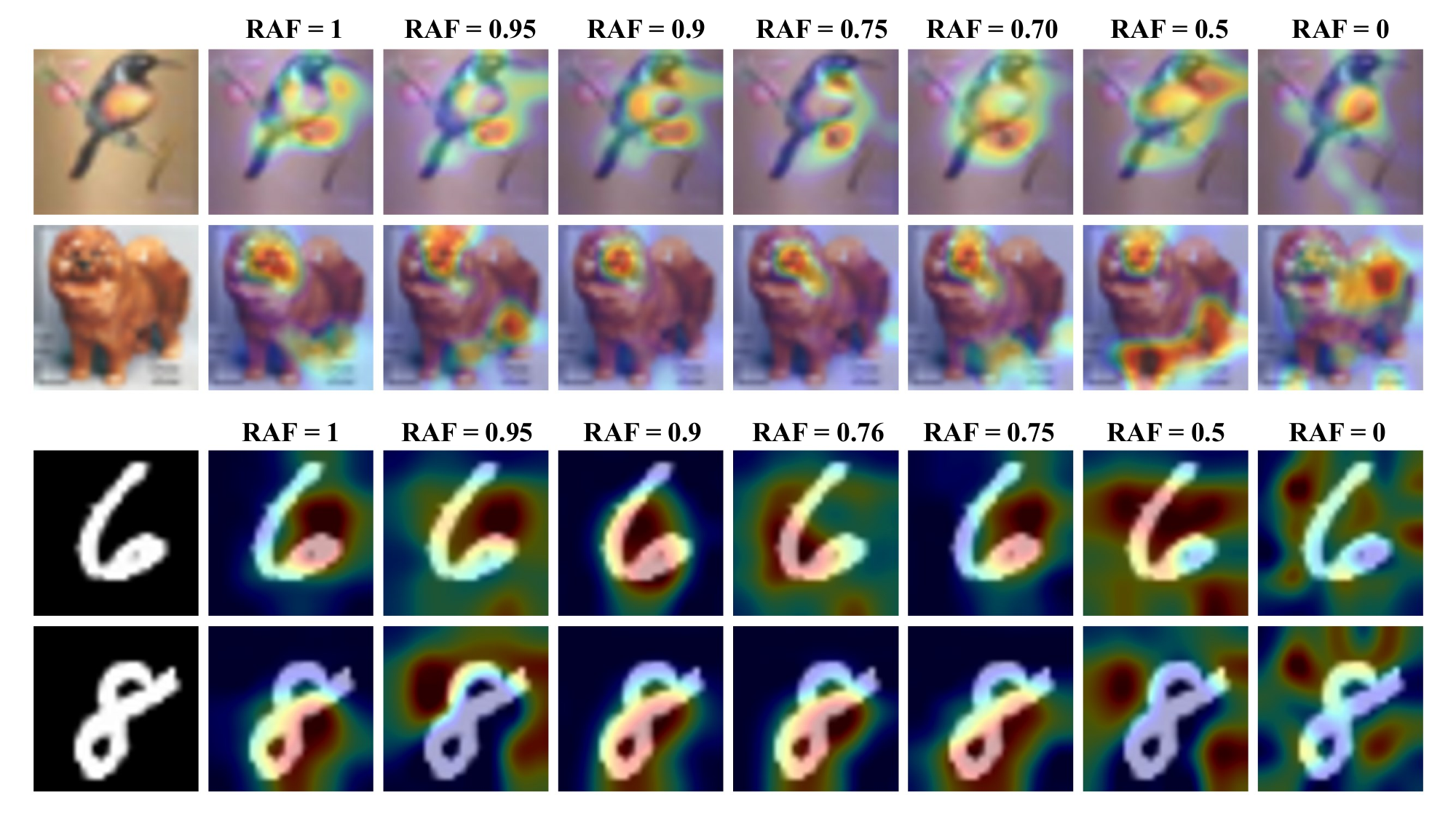}
  \caption{\textbf{Grad-CAM Visualization Results of the Last Residual Block's Feature Map under ResNet-50 Configuration for MNIST (bottom) and CIFAR-10 (top).} As \( RAF \) decreases within a specified range, the model focuses more on the contours of the digits and is able to extract details with greater precision.}
  \label{fig_4}
\end{figure}

Fig.~\ref{fig_4} visualizes this effect for ResNet‑50.  At \( RAF = 0.95 \) the network, when classifying MNIST digits, redirects its focus from large uniform strokes to finer terminators and junctions, lifting accuracy from 99.28\% to 99.36\%.  
Further lowering \( RAF \) to 0.75 during CIFAR‑10 training accentuates biologically meaningful parts—eyes, muzzles, and wing tips—yielding a 0.48‑point gain (90.77\% vs.\ 90.29\%).  
Below \( RAF = 0.70 \), however, the heat‑maps fragment into multiple low‑energy blobs, signalling over‑abstraction and a concomitant 8–10\% drop in top‑1 accuracy.
\subsubsection{ShuffleNet v2.}  
Fig.~\ref{fig_5} shows analogous behaviour for the lightweight ShuffleNet v2.  A mild cut to \( RAF = 0.95 \) sharpens digit contours on MNIST (99.51\% vs.\ 99.35\%), while \( RAF = 0.80 \) on CIFAR‑10 steers attention toward species‑defining shapes such as canine snouts or avian beaks, adding 0.40 pp over baseline.  Excessive reduction (\( RAF = 0.50 \)) again leads to scattered saliency, mirroring the accuracy collapse seen in ResNet‑50.

These qualitative heat‑map shifts and their quantitative echoes corroborate our core hypothesis: \( RAF \) acts as a controllable valve between global classification objectives and localized feature convergence.  When tuned within the empirically determined window of 0.75–0.95, it encourages the discovery of physically interpretable patterns while preserving, and often enhancing, overall performance.

\section{Conclusion}
This work demonstrates the effectiveness of evaluating feature map convergence through the proposed FMCE-Net++ framework, which integrates auxiliary supervision into neural network training. By adaptively combining feature map convergence scores with traditional loss functions using a Representation Abstraction Factor ($RAF = 0.70 - 0.95$), our method achieves notable performance gains, including a 1.16\% accuracy improvement on CIFAR-10 with ResNet-50 and a 0.40\% improvement with ShuffleNet v2. The architecture-independent effectiveness of the proposed method is confirmed through Grad-CAM visualizations, indicating enhanced feature localization capabilities. Although the current implementation employs basic evaluation networks, future efforts will focus on further optimizing the FMCE-Net architecture to broaden its applicability. Overall, this approach introduces promising opportunities for module-level optimization in deep learning frameworks, and significant potential remains for further exploring feature map convergence evaluation methods.

\section*{Acknowledgments}
This work was supported by the National Key R\&D Program of China: ``Development of Large Model Technology and Scenario Library Construction for Autonomous Driving Data Closed-Loop'' (Grant No. 2024YFB2505501); the Guangxi Key Science and Technology Project: ``Research and Industrialization of High-Performance and Cost-Effective Urban Pilot Driving Technologies'' (Grant No. AA24206054); and the Independent Research Project of the State Key Laboratory of Intelligent Green Vehicle and Mobility, Tsinghua University (Grant No. ZZ-GG-20250405).

\bibliographystyle{IEEEtran}
\bibliography{reference}

\end{document}